\documentclass[10pt,twocolumn,letterpaper]{article}

\usepackage[pagenumbers]{cvpr} 

\usepackage[accsupp]{axessibility}  

\usepackage{adjustbox}
\usepackage{multirow}
\usepackage{makecell}
\usepackage{amssymb}
\usepackage{pifont}
\newcommand{\cmark}{\ding{51}}%
\newcommand{\xmark}{\ding{55}}%

\definecolor{cvprblue}{rgb}{0.21,0.49,0.74}
\usepackage[pagebackref,breaklinks,colorlinks,allcolors=cvprblue]{hyperref}

\def\paperID{1} 
\def\confName{CVPR}
\def\confYear{2025}

\title{Strong Baseline: Multi-UAV Tracking via YOLOv12 with BoT-SORT-ReID}

\author{Yu-Hsi Chen \\
The University of Melbourne\\
Parkville, Australia\\
{\tt\small yuhsi@student.unimelb.edu.au}
}

\begin{document}
\maketitle
\begin{abstract}
Detecting and tracking multiple unmanned aerial vehicles (UAVs) in thermal infrared video is inherently challenging due to low contrast, environmental noise, and small target sizes. This paper provides a straightforward approach to address multi-UAV tracking in thermal infrared video, leveraging recent advances in detection and tracking. Instead of relying on the well-established YOLOv5 with DeepSORT combination, we present a tracking framework built on YOLOv12 and BoT-SORT, enhanced with tailored training and inference strategies. We evaluate our approach following the 4th Anti-UAV Challenge metrics and reach competitive performance. Notably, we achieved strong results without using contrast enhancement or temporal information fusion to enrich UAV features, highlighting our approach as a "Strong Baseline" for multi-UAV tracking tasks. We provide implementation details, in-depth experimental analysis, and a discussion of potential improvements.
The code is available at \url{https://github.com/wish44165/YOLOv12-BoT-SORT-ReID}.
\end{abstract}

\section{Introduction}
\label{sec:intro}

Multi-UAV tracking has emerged as a crucial application in recent years, driven by significant advancements in hardware, detection models, and tracking algorithms. As UAVs equipped with sophisticated visual systems and advanced control dynamics continue to proliferate, a wide range of UAV-based products have been introduced, as presented in~\cite{nyboe2022mpsoc4drones}. However, these innovations also introduce new challenges, particularly in tracking UAV swarms. The need for effective swarm tracking has become increasingly urgent due to growing security concerns and the rising threat posed by unauthorized UAVs. Various UAV-related datasets have been developed to address these challenges to advance tracking and detection tasks. These datasets include trajectory reconstruction datasets, such as those in~\cite{li2020reconstruction,hwang20233d}, which provide UAV trajectories captured from single or multi-view cameras, and trajectory-based UAV datasets introduced in~\cite{chen2024uavdb}.
Additionally, RGB-based footage datasets, including those in~\cite{pawelczyk2020real,steininger2021aircraft,li2016multi,Airborne_Object_Tracking_AWS}, have been widely used. Among these, thermal infrared video-based UAV datasets featuring both single-object tracking (SOT) and multi-object tracking (MOT) scenarios in~\cite{jiang2021anti,huang2023anti}, have gained significant attention, particularly in major challenge events. These datasets have played a pivotal role in improving UAV tracking and detection capabilities.

\begin{figure*}[t]
    \centering
    \includegraphics[width=\linewidth]{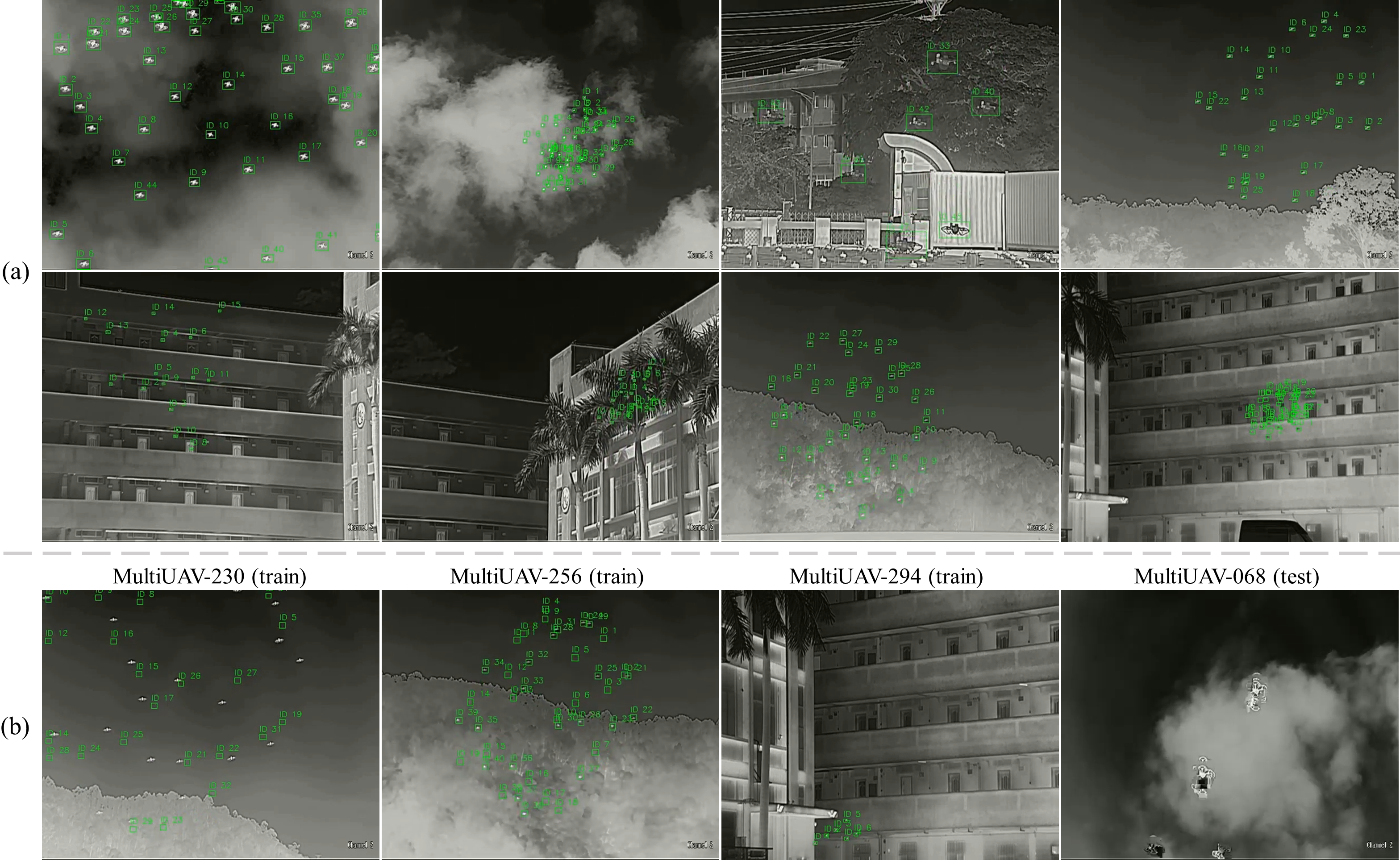}
    \caption{Demonstration using training and testing data from Track 3. (a) Shows UAV swarms with varying sizes and backgrounds in the training data. (b) Highlights annotation errors and frame defects: MultiUAV-230 (train) has incorrect annotations, MultiUAV-256 (train) contains redundant annotations, MultiUAV-294 (train) has missed annotations, and MultiUAV-068 (test) includes a poor-quality frame.}
    \label{fig:data}
\end{figure*}

\begin{figure*}[t]
    \centering
    \includegraphics[width=\linewidth]{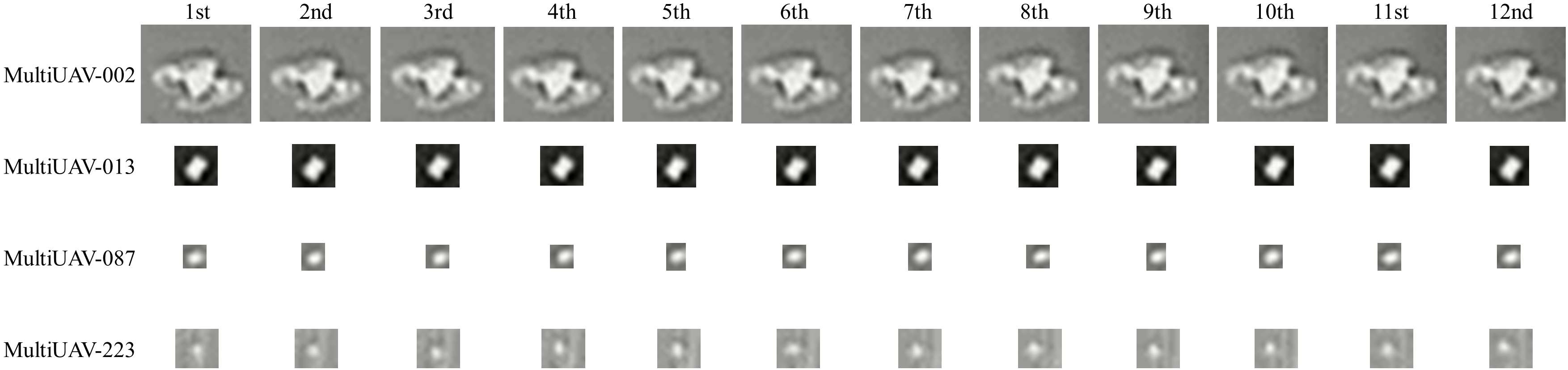}
    \caption{Illustration of cropped image patches from training data annotations. Each patch corresponds to a bounding box from MultiUAV-002, MultiUAV-013, MultiUAV-087, and MultiUAV-223, with sizes approximately \( 28 \times 24 \), \( 10 \times 10 \), \( 6 \times 6 \), and \( 11 \times 10 \), respectively. The top number denotes the frame sequence (1st to 12th frames), and each row represents the same object (same ID across frames).}
    \label{fig:reid}
\end{figure*}

Thermal infrared videos offer advantages over traditional RGB imagery, such as enhanced visibility in low-light and adverse weather conditions, making them ideal for security and surveillance applications. This paper focuses on using thermal infrared video for multi-UAV tracking, exploiting its importance in challenging environments where RGB-based methods may fail. \cref{fig:data} (a) illustrates thermal infrared frames with diverse backgrounds from the MOT training set, while \cref{fig:data} (b) highlights minor defects, such as annotation errors, redundancies, missed labels, and low-quality frames, that account for a negligible portion of the dataset and can be safely disregarded during training. Additionally, \cref{fig:reid} displays cropped image patches from bounding box annotations in the training set, illustrating the varying sizes of UAVs, from several pixels to single-digit pixels.
We build a complete UAV tracking workflow by leveraging the latest YOLOv12~\cite{tian2025yolov12} detector and BoT-SORT~\cite{aharon2022bot} tracking algorithm, which outperform the YOLOv5~\cite{jocher2020ultralytics} with the DeepSORT~\cite{wojke2017simple} pipeline. We also implement some strategies to enhance multi-UAV tracking performance further. Our contributions are as follows:
\begin{itemize}
    \item[1.] We establish a multi-UAV tracking workflow based on YOLOv12 and BoT-SORT, setting a strong baseline for thermal infrared video-based multi-UAV tracking tasks.
    \item[2.] We provide insightful analysis of various trial adjustments, such as the impact of input image size and tracker buffer tuning, and offer essential considerations for future improvements starting from our strong baseline.
\end{itemize}

\section{Related Work}
\label{sec:related_work}

Existing perspectives for improving thermal infrared video-based multi-UAV tracking can be categorized into annotation and benchmarking, spatial information enhancement, temporal and motion modeling, real-time optimization, unified frameworks, and detection-based tracking systems.
As high-quality annotation is fundamental for robust tracking, prior studies have examined the impact of annotation errors on object detection~\cite{koksal2020effect}, incorporated Multiple Hypothesis Tracking (MHT) to leverage temporal cues and reduce false positives~\cite{ince2021semi}, and introduced benchmarks to evaluate detection and tracking methods on UAV datasets~\cite{isaac2021unmanned}. Spatial information enhancement techniques, such as the Image Pyramid Guidance (IPG) module presented in~\cite{liu2020ipg}, address feature imbalance by preserving fine-grained spatial details for accurate bounding box regression and classification, even in deep network layers.

To further improve tracking robustness, temporal and motion modeling techniques exploit inter-frame correlations, enhancing continuity and reducing fragmentation~\cite{huang2021siamsta,he2023motion,li2023global,yang2023video}. Complementary to this, real-time optimization strategies reduce inference latency while maintaining accuracy, enabling efficient UAV tracking in real-world applications~\cite{wang2020real,wu2020real,fang2021real,lyu2023real}.  
Beyond these, unified frameworks integrate detection and tracking into end-to-end solutions, streamlining multi-UAV tracking pipelines~\cite{zhao2021unified,yu2023unified}. Additionally, detection-based methods incorporating cascading post-processing modules refine tracking accuracy by mitigating false positives and improving localization~\cite{tang2023strong}.  
While prior works have contributed significantly to multi-UAV tracking, our approach advances the field by leveraging the latest detector and tracker, setting a new benchmark for thermal infrared video-based UAV tracking and guiding future research in the multi-UAV tracking task.
\section{Methodology}
\label{sec:methodology}

This section first defines the problem scope, followed by data analysis and preparation for model training. We then introduce the primary detection model, YOLOv12, and the tracking algorithm, BoT-SORT, before detailing our training and inference strategies.

\subsection{Problem Statement}
\label{ssec:problem_statment}

The goal is to track UAVs as accurately as possible, with evaluation metrics detailed in \cref{ssec:evaluation}. The challenge consists of three tracks, each corresponding to a different scenario. Tracks 1 and 2 are SOT tasks, differing in whether the UAV's initial location is given. Track 3 is a MOT task where the initial locations of UAVs are provided.

\subsection{Data Analysis and Preparation}
\label{ssec:analysis_preprocess}

We first analyze each track's training and testing data, as summarized in \cref{tab:data_characteristics}. Tracks 1 and 2 share the same training set, consisting of 23 sequences at 512\(\times\)512 resolution and 200 sequences at 640\(\times\)512 resolution. The training set for Track 3 is composed of 200 sequences at 640\(\times\)512 resolution.  
For testing, Tracks 1 and 2 each contain 216 non-overlapping sequences. Track 1's test set is entirely at 640\(\times\)512 resolution, while Track 2 includes 16 sequences at 640\(\times\)512 and 200 at 512\(\times\)512. Track 3's test set consists of 100 sequences at 640\(\times\)512 resolution.
Additionally, \cref{tab:data_characteristics} reports the width, height, and area distributions, along with their mean and standard deviation, providing essential insights for model hyperparameter tuning. Note that there may be slight differences in the numbers compared to the official release, as we have removed redundant annotations and defect cases, as illustrated in \cref{fig:data} (b).

\begin{table*}
    \centering
    \begin{adjustbox}{max width=\linewidth}
    \begin{tabular}{@{}lcccccc@{}}
        \toprule
        \multirow{2.5}{*}{Characteristic} & \multicolumn{3}{c}{Training Data} & \multicolumn{3}{c}{Testing Data}\\
        \cmidrule(r){2-4} \cmidrule(l){5-7}
        & Track 1 & Track 2 & Track 3 & Track 1 & Track 2 & Track 3 \\
        \midrule
        Number of Sequences & \multicolumn{2}{c}{23 / 200} & 200 & 216 & 16 / 200 & 100\\
        Number of Frames & \multicolumn{2}{c}{16,022 / 231,557} & 151,831 & 232,742 & 11,619 / 221,123 & 75,487 \\
        Resolutions & \multicolumn{2}{c}{512\(\times\)512 / 640\(\times\)512} & 640\(\times\)512 & 640\(\times\)512 & 512\(\times\)512 / 640\(\times\)512 & 640\(\times\)512\\
        Total Bounding Boxes & \multicolumn{2}{c}{229,839} & 3,127,045 & 216 & 0 & 2,041 \\
        Width Range (px) & \multicolumn{2}{c}{[1, 146]} & [0.96, 98.92] & [4, 140] & N/A & [1.86, 28.33]\\
        Width Mean \(\pm\) Std (px) & \multicolumn{2}{c}{30.55 \(\pm\) 24.43} & 10.56 \(\pm\) 5.75 & 40.56 \(\pm\) 26.34 & N/A & 9.71 \(\pm\) 4.06 \\
        Height Range (px) & \multicolumn{2}{c}{[1, 131]} & [0.57, 55.5] & [3, 68] & N/A & [2.44, 25.78]\\
        Height Mean \(\pm\) Std (px) & \multicolumn{2}{c}{19.43 \(\pm\) 13.63} & 9.06 \(\pm\) 4.67 & 23.76 \(\pm\) 13.34 & N/A & 8.56 \(\pm\) 3.55\\
        Area Range (px\(^2\)) & \multicolumn{2}{c}{[1, 17,161]} & [0.95, 4344.63] & [16, 7,956] & N/A & [10.88, 575.41]\\
        Area Mean \(\pm\) Std (px\(^2\)) & \multicolumn{2}{c}{874.91 \(\pm\) 1158.50} & 119.05 \(\pm\) 179.40 & 1241.12 \(\pm\) 1280.20 & N/A & 95.14 \(\pm\) 86.91\\
        \bottomrule
    \end{tabular}
    \end{adjustbox}
    \caption{Summary of data characteristics for training and testing. The training data for Tracks 1 and 2 are identical, while the testing data for Tracks 1 and 2 have no overlapping parts. Since Track 2 does not provide initial bounding boxes, we use "N/A" to indicate that this information is not applicable. Note that some bounding boxes were removed in Tracks 1 and 2 training sets due to labeled non-existent, zero-sized width or height, and cases where the boxes covered the entire image.}
    \label{tab:data_characteristics}
\end{table*}

After analyzing the data, we split it for model training preparation. The number of frames and bounding boxes used for training, validation, and testing in the SOT and MOT tasks are detailed in \cref{tab:data_preparation}. Specifically, Tracks 1 and 2 use YOLOv12 with BoT-SORT, while Track 3 employs YOLOv12 with BoT-SORT-ReID. Note that some numbers are in parentheses since we found the test set to provide limited information for the SOT task. Thus, the values in parentheses reflect the data only split into training and validation sets. Additionally, for BoT-SORT training, 1/10 of the data is primarily used to train the ReID module. This approach provides more effective ReID module training since many scenes are visually similar.

\begin{table*}
    \centering
    \begin{adjustbox}{max width=\linewidth}
    \begin{tabular}{@{}lccccccccc@{}}
        \toprule
        \multirow{3.5}{*}{Characteristic} & \multicolumn{3}{c}{Single-Object Tracking} & \multicolumn{4}{c}{Multi-Object Tracking}\\
        \cmidrule(r){2-4} \cmidrule(l){5-8}
        & \multicolumn{3}{c}{YOLOv12} & \multicolumn{2}{c}{YOLOv12} & \multicolumn{2}{c}{BoT-SORT-ReID}\\
        \cmidrule(r){2-4} \cmidrule(lr){5-6} \cmidrule(l){7-8}
        & Train & Valid & Test & Train & Valid & Train & Valid\\
        \midrule
        Number of Frames & 148,547 (198,063) & 49,515 (49,516) & 49,517 & 121,355 & 30,337 & 75,913 (7,593) & 75,918 (7,783) \\
        Total Bounding Boxes & 138,084 (184,056) & 45,848 (45,921) & 46,045 & 2,501,753 & 625,292 & 1,580,931 (155,833) & 1,546,092 (160,876) \\
        \bottomrule
    \end{tabular}
    \end{adjustbox}
    \caption{Data preparation summary for YOLOv12 and BoT-SORT-ReID across single-object (Tracks 1 and 2) and multi-object (Track 3) tracking. For YOLOv12 in SOT, numbers in parentheses indicate an alternative split with only training and validation sets. For BoT-SORT-ReID, numbers in parentheses indicate a reduced ReID training set selecting only the first 1/10 of frames from each sequence.}
    \label{tab:data_preparation}
\end{table*}

\subsection{YOLOv12 with BoT-SORT-ReID for MOT}
\label{ssec:yolov12_botsort}

Based on the comprehensive evaluation results presented in~\cite{chen2024uavdb}, which benchmarks the YOLO series of detectors on UAV datasets featuring RGB footage, YOLOv12 was selected for all tracks due to its superior performance. YOLOv12~\cite{tian2025yolov12} represents the latest advancement in the YOLO series of object detectors, introducing key innovations to enhance accuracy and efficiency simultaneously. At its core, YOLOv12 adopts the Residual Efficient Layer Aggregation Network (R-ELAN), which addresses the optimization challenges associated with attention mechanisms, particularly in large-scale models. Building upon ELAN~\cite{wang2023yolov7}, R-ELAN introduces a block-level residual design with adaptive scaling alongside a refined feature aggregation strategy, jointly promoting effective feature reuse and stable gradient propagation with minimal overhead. Furthermore, YOLOv12 integrates an attention-centric architecture by combining FlashAttention~\cite{dao2022flashattention,dao2023flashattention} with spatially aware modules, enabling enhanced contextual modeling while preserving low latency. Introducing 7\(\times\)7 large-kernel separable convolutions broadens the receptive field and strengthens object localization, particularly for small and medium-sized targets. The architecture is optimized for modern GPU memory hierarchies, delivering improved computational efficiency and reduced inference times without compromising detection performance. These innovations enable YOLOv12 to balance speed and accuracy, making it highly suitable for real-time applications, large-scale detection tasks, and tracking pipelines.

BoT-SORT~\cite{aharon2022bot} combines a Kalman Filter~\cite{wojke2017simple} with camera motion compensation (CMC) to stabilize tracking under dynamic conditions. CMC employs global motion compensation (GMC) via affine transformations, using image keypoints~\cite{shi1994good} tracked with pyramidal Lucas-Kanade optical flow~\cite{bouguet2001pyramidal} and outlier rejection. The affine transformation, estimated via RANSAC~\cite{fischler1981random}, compensates for background motion while maintaining object trajectory stability by adjusting Kalman Filter state vectors.
BoT-SORT-ReID enhances multi-object tracking by integrating appearance cues from four distinct ReID architectures. The Bag of Tricks (Bagtricks) baseline employs a ResNet-50 backbone with batch normalization, triplet loss, and cross-entropy loss for robust feature extraction. Attention Generalized-Mean Pooling with Weighted Triplet Loss (AGW)~\cite{ye2021deep} improves feature representation by incorporating non-local modules and generalized mean pooling. Strong Baseline (SBS)~\cite{luo2019bag} enhances robustness with generalized mean pooling, circle softmax loss, and an advanced data augmentation strategy. Multiple Granularity Network (MGN)~\cite{wang2018learning} extends SBS by introducing multiple feature branches to capture fine-grained representations across different spatial scales. Additionally, linear tracklet interpolation with a 20-frame gap, following ByteTrack~\cite{zhang2022bytetrack}, mitigates missed detections from occlusions or annotation errors.

\subsection{Training and Inference Strategies} 
\label{ssec:strategies}

To reduce the training time of the YOLOv12 detector, we adopt a two-stage training strategy. First, we train YOLOv12 models (n, s, m, l, x) from scratch on the SOT dataset, which is split into training, validation, and testing subsets as detailed in \cref{tab:data_preparation}. Subsequently, starting from this checkpoint, we fine-tune these models on the MOT dataset or with larger input image resolutions. This staged approach accelerates convergence, reduces overall training time, and enables the model to achieve competitive Average Precision (AP) within just a few epochs. For the ReID module, we primarily employ a reduced subset of the dataset to enhance training efficiency, as using the entire dataset for training would be highly time-consuming.

\begin{figure*}[t]
    \centering
    \includegraphics[width=\linewidth]{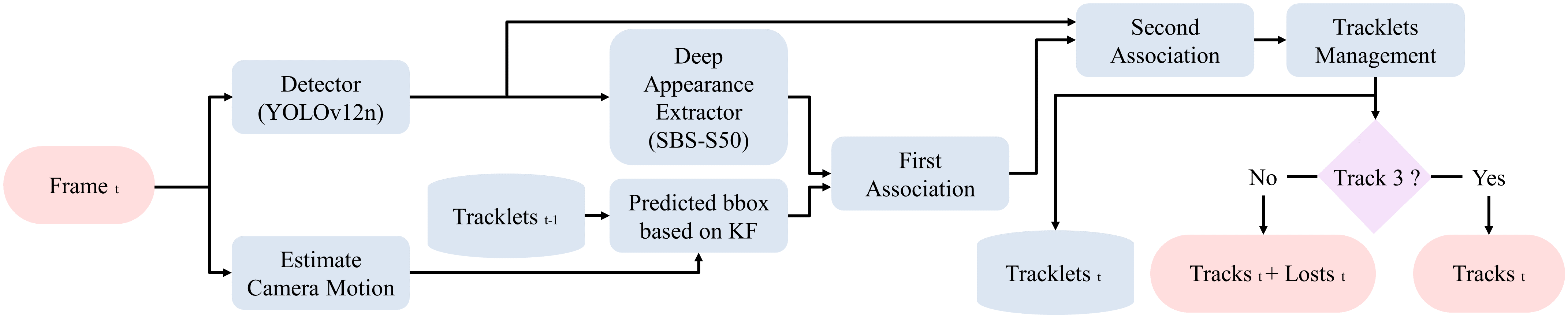}
    \caption{YOLOv12n with BoT-SORT-SBS-S50 workflow diagram. The workflow follows the original BoT-SORT framework~\cite{aharon2022bot}, with a slight revision: incorporating lost tracks to compensate for uninformative frames and improve object continuity. Specifically, for Tracks 1 and 2, lost target information is used to annotate potential object locations, while Track 3 retains the BoT-SORT original output.}
    \label{fig:workflow}
\end{figure*}

The inference workflow is presented in \cref{fig:workflow}. The overall procedure follows the original BoT-SORT scheme. However, we modify the output by reporting both online and lost targets for Tracks 1 and 2 while preserving the original output format for Track 3. We did not use linear track interpolation because ID switching frequently occurs due to camera motion or fast-moving UAVs, making interpolation ineffective for recovering missing detections. Instead, for the SOT task, we adopt a strategy based on the assumption that each frame contains at most one UAV, following this priority order: (1) report the UAV with the highest confidence score among the online targets, (2) if no online target is available, continue reporting the previous ID as the lost target in the subsequent tracker buffer frames, (3) if no previous ID is available, report the last known location until new online targets are detected. This strategy leverages the Kalman Filter's prediction to accurately estimate the UAV's location based on prior positions and velocity, significantly improving evaluation metrics in the SOT task. However, this strategy is not feasible for the MOT task due to the frequent overlap and ID switching between online and lost targets, which would lead to poor results. Therefore, we maintain the original output for Track 3 in this case.

\section{Experimental Results}
\label{sec:results}

The experiments were conducted on two platforms: (1) a local system equipped with an Intel Core i7-12650H CPU, NVIDIA RTX 4050 GPU, and 16 GB of RAM, primarily used for data processing, and (2) a high-performance computing (HPC) system~\cite{meade2017spartan} featuring an NVIDIA H100 GPU and 80 GB of memory, used mainly for model training. All models were trained using the default settings (\eg, image input size of 640 and a track buffer of 30 frames) unless otherwise specified in the content or tables. This section begins by outlining the evaluation metrics for the three tracks, followed by the results for both the SOT and MOT tasks. We then present the leaderboard rankings and discuss key considerations and potential image enhancement techniques that could further improve UAV tracking.

\subsection{Evaluation Metrics}
\label{ssec:evaluation}

Two evaluation metrics are used across the three competition tracks. The first metric applies to Tracks 1 and 2, where tracking accuracy is defined as:  
\begin{equation}
\begin{aligned}
    \text{acc} &= \sum\limits_{t=1}^T \frac{\text{IoU}_t \cdot \delta (v_t>0) + p_t \cdot (1 - \delta(v_t>0))}{T} \\
    & \hspace*{4mm} - 0.2 \cdot \left(\sum\limits_{t=1}^{T^*} \frac{p_t \cdot \delta(v_t>0)}{T^*}\right)^{0.3}.
    \label{eq:eval_sot}
\end{aligned}    
\end{equation}  
Here, \( \text{IoU}_t \) is the Intersection over Union between the predicted and ground-truth bounding boxes at frame \( t \). The variable \( p_t \) denotes the predicted visibility flag, where \( p_t = 1 \) if the predicted box is empty, and \( p_t = 0 \) otherwise. The ground-truth visibility is given by \( v_t \), and \( \delta(v_t > 0) \) is an indicator function that equals 1 when the target is visible (\( v_t > 0 \)) and zero otherwise. The accuracy is averaged over all \( T \) frames, with \( T^* \) representing the number of frames in which the target is visible in the ground truth. The second metric used for Track 3 is the Multi-Object Tracking Accuracy (MOTA), which jointly penalizes false positives (FP), false negatives (FN), and identity switches (IDS), normalized by the total number of ground-truth objects (GT):  
\begin{equation}
    \text{MOTA} = 1 - \frac{\text{FP} + \text{FN} + \text{IDS}}{\text{GT}}.
    \label{eq:eval_mot}
\end{equation}  
MOTA ranges from \( -\infty \) to 1, with higher values indicating better tracking performance. The final score is obtained by averaging MOTA over all sequences. The following sections will present and evaluate all performance results based on the abovementioned metrics.

\subsection{Evaluation Results on Tracks 1 and 2}
\label{ssec:results_track1_track2}

\begin{table*}
    \centering
    \begin{adjustbox}{max width=\linewidth}
    \begin{tabular}{@{}lcccccccccccccccc@{}}
    \toprule
    \multirow{2.5}{*}{Configurations}& \multicolumn{8}{c}{Track 1} & \multicolumn{8}{c}{Track 2}\\
    \cmidrule(lr){2-9} \cmidrule(l){10-17}
    & Trial 1 & Trial 2 & Trial 3 & Trial 4 & Trial 5 & Trial 6 & Trial 7 & Trial 8 & Trial 1 & Trial 2 & Trial 3 & Trial 4 & Trial 5 & Trial 6 & Trial 7 & Trial 8 \\
    \midrule
    YOLOv12 & n & n & s & m & l & x & x & n & n & n & s & m & l & x & x & n\\
    - epochs & 100 & 100 & 100 & 100 & 100 & 100 & 300 & 100 & 100 & 100 & 100 & 100 & 100 & 100 & 300 & 100 \\
    BoT-SORT & \xmark & \cmark & \cmark & \cmark & \cmark & \cmark & \cmark & \cmark & \xmark & \cmark & \cmark & \cmark & \cmark & \cmark & \cmark & \cmark \\
    - min box area & 10 & 10 & 10 & 10 & 10 & 10 & 10 & 4 & 10 & 10 & 10 & 10 & 10 & 10 & 10 & 1 \\
    Scores & 0.0786 & 0.5529 & 0.5637 & 0.5634 & 0.5644 & 0.5548 & 0.5398 & \textbf{0.5813} & 0.0992 & 0.3106 & 0.3258 & 0.3283 & 0.3285 & 0.3132 & 0.3080 & \textbf{0.3559} \\
    \bottomrule
    \end{tabular}
    \end{adjustbox}
    \caption{Evaluation results for Tracks 1 and 2, summarizing eight trials per track. The first two rows detail YOLOv12 configurations, varying model sizes (n, s, m, l, x), and training epochs. The third row specifies the use of BoT-SORT ("\xmark" indicates exclusion; "\cmark" indicates inclusion). The fourth row lists the "min box area" threshold for filtering tiny bounding boxes, adjusted to 4 and 1 to account for tiny UAVs. The final row reports the resulting scores.}
    \label{tab:results_track1_track2}
\end{table*}

We present the evaluation results for Tracks 1 and 2 together, as both are SOT tasks, with the only difference being the presence of the initial UAV location. Eight meaningful trials are selected for both tracks, as shown in \cref{tab:results_track1_track2}. Trials 1 and 2 serve as an ablation study to assess the impact of BoT-SORT. The results demonstrate a significant performance improvement: the score in Track 1 increases from 0.0786 to 0.5529, and in Track 2, it rises from 0.0992 to 0.3106 simply by adding BoT-SORT after the YOLOv12n detector. Trials 2 through 6 evaluate different detector model sizes (n, s, m, l, x), with the highest score achieved using YOLOv12l for both tracks. Trial 7 examines the effect of extended 300 epochs training, revealing a decline in performance compared to the 100 epochs training, likely due to overfitting. Finally, Trial 8 for each track shows the highest score we submitted, tuning the minimum box area threshold from 10 to 4 for Track 1 and from 10 to 1 for Track 2 to better capture smaller UAVs that may have been missed with the default setting.

\subsection{Evaluation Results on Track 3}
\label{ssec:results_track3}

\begin{table*}
    \centering
    \begin{adjustbox}{max width=\linewidth}
    \begin{tabular}{@{}lcccccccccc@{}}
    \toprule
    \multirow{2.5}{*}{Configurations} & \multicolumn{5}{c}{Group 1} & \multicolumn{5}{c}{Group 2} \\
    \cmidrule(lr){2-6} \cmidrule(l){7-11}
    & Trial 1 & Trial 2 & Trial 3 & Trial 4 & Trial 5 & Trial 6 & Trial 7 & Trial 8 & Trial 9 & Trial 10 \\
    \midrule
    YOLOv12 & n & s & m & l & x & n & n & n & n & n \\
    BoT-SORT & \cmark & \cmark & \cmark & \cmark & \cmark & \cmark & \cmark & \cmark & \cmark & \cmark \\
    - track buffer & 30 & 30 & 30 & 30 & 30 & 15 & 45 & 60 & 75 & 90 \\
    Scores & 0.638763 & 0.635361 & 0.633887 & 0.631864 & 0.630822 & 0.638609 & 0.638781 & 0.638801 & 0.638788 & 0.638771\\
    \midrule 
    \multirow{2.5}{*}{Configurations} & \multicolumn{2}{c}{Group 3} & \multicolumn{7}{c}{Group 4} & Final \\
    \cmidrule(lr){2-3} \cmidrule(lr){4-10} \cmidrule(l){11-11}
    & Trial 11 & Trial 12 & Trial 13 & Trial 14 & Trial 15 & Trial 16 & Trial 17 & Trial 18 & Trial 19 & Trial 20 \\
    \midrule
    YOLOv12 & n & n & n & n & n & n & n & n & n & n \\
    - image size & 1280 & 1600 & 640 & 640 & 640 & 640 & 640 & 640 & 640 & 1600 \\
    - epochs & 42 & 20 & 100 & 100 & 100 & 100 & 100 & 100 & 100 & 11 \\
    BoT-SORT & \cmark & \cmark & \cmark & \cmark & \cmark & \cmark & \cmark & \cmark & \cmark & \cmark \\
    - track buffer & 60 & 60 & 30 & 30 & 30 & 30 & 30 & 30 & 30 & 60 \\
    - ReID module & \xmark & \xmark & sbs\_S50 & sbs\_R101-ibn & sbs\_S50 & sbs\_S50 & sbs\_S50 & sbs\_S50 & sbs\_S50 & sbs\_S50 \\
    - metric learning & N/A & N/A & TripletLoss & TripletLoss & TripletLoss & CircleLoss & CircleLoss & CircleLoss & CircleLoss & CircleLoss\\
    - epochs & N/A & N/A & 8 & 58 & 60 & 120 & 60 & 33 & 17 & 17 \\
    Scores & 0.749352 & 0.744046 & 0.647239 & 0.646299 & 0.647056 & 0.647290 & 0.647423 & 0.647567 & 0.647591 & \textbf{0.760874} \\
    \bottomrule
    \end{tabular}
    \end{adjustbox}
    \caption{Comprehensive evaluation results for Track 3. Experiments are grouped into four categories: (1) varying YOLOv12 model sizes, (2) tuning BoT-SORT's track buffer, (3) exploring input image resolutions, and (4) configuring ReID modules and training strategies. The final configuration, guided by these studies, achieves the highest score, confirming the effectiveness of our optimization.}
    \label{tab:results_track3}
\end{table*}

The evaluation results for Track 3 can be categorized into four key observations. As shown in \cref{tab:results_track3}, Group 1 presents results using various YOLOv12 model sizes, revealing that YOLOv12n achieves the best performance despite being the smallest model. Group 2 examines the effect of different track buffer sizes, with the highest score observed using 60 buffer frames, suggesting that this configuration optimizes the ID reassociation process. Group 3 investigates the impact of varying image input sizes. Both 1280 and 1600 input sizes, compared to the default 640, result in a significant performance boost. Group 4 discusses trials involving different ReID modules. Trial 13 uses the full ReID dataset, while Trials 14 through 19 are trained on a reduced ReID dataset. This group also evaluates the influence of different configurations, including changes in the ReID module structure, metric learning strategies, and the number of training epochs. From these results, we draw the following conclusions: (1) ResNet-50 from the Strong Baseline Series outperforms ResNet-101 with Instance-Batch Normalization as the backbone for the ReID module, (2) replacing Triplet Loss with CircleLoss for metric learning leads to improved performance, and (3) ReID module training tends to overfit as the number of epochs increases.

Based on all trials across the groups, we draw the following conclusions regarding score variations relative to Trial 1: (1) Model size affects performance by approximately 0.001, (2) Track buffer size influences the score by around 0.0001, (3) Image input size contributes the most significant impact, with a score increase of about 0.1, and (4) the ReID module accounts for roughly 0.01. Leveraging these insights, Trial 20, which achieved the highest score we submitted, adopts the following configuration: YOLOv12n with an image size of 1600, trained for 11 epochs, combined with BoT-SORT-SBS-S50 equipped with CircleLoss, optimized with AdamW~\cite{loshchilov2019decoupledweightdecayregularization} and trained for 17 epochs.

\begin{figure*}[t]
    \centering
    \includegraphics[width=\linewidth]{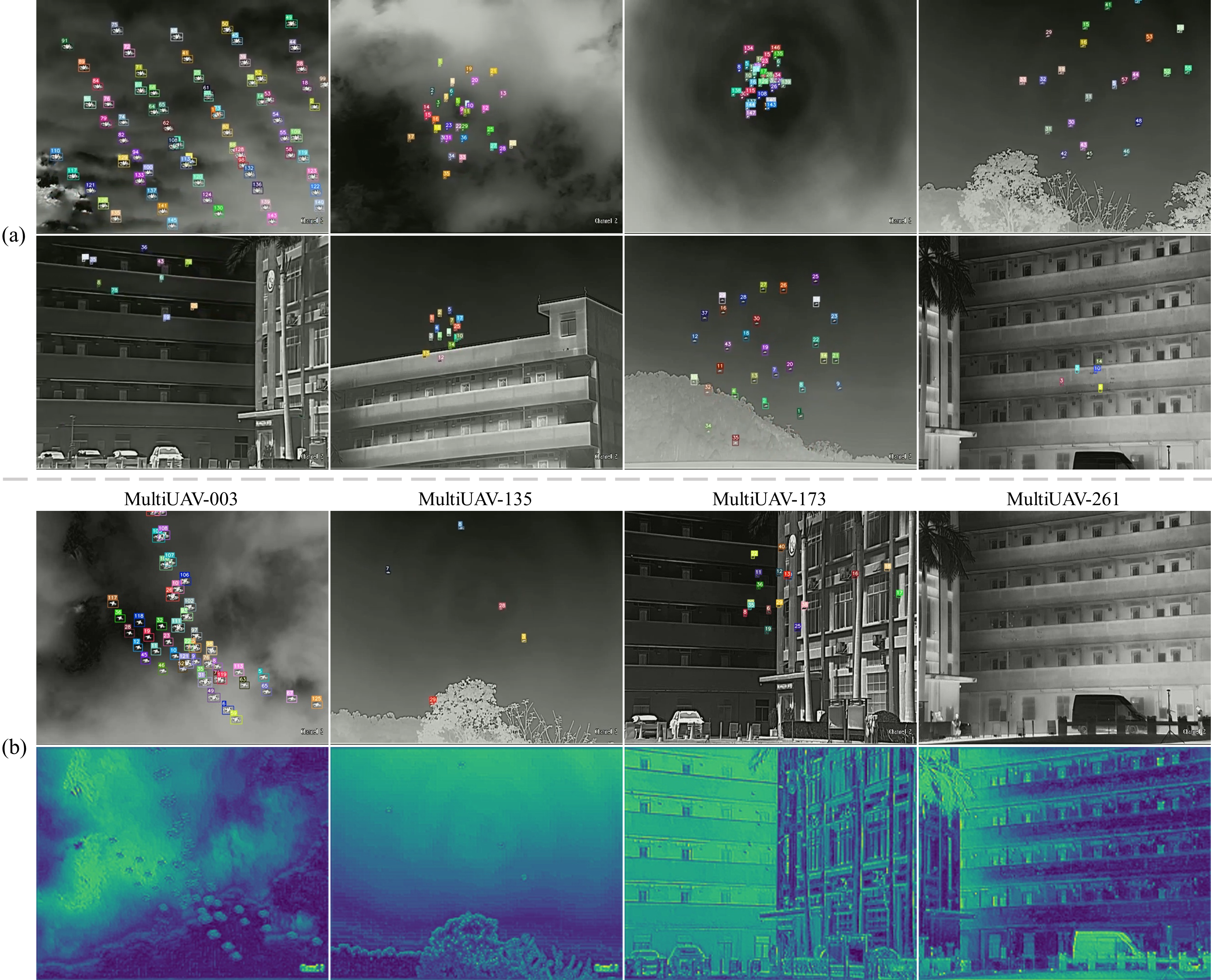}
    \caption{Demonstration of YOLOv12n with BoT-SORT-SBS-S50 predictions on Track 3 test data. (a) Predicted bounding boxes with object IDs. (b) Challenging scenarios: MultiUAV-0003 contains multiple overlapping UAVs; MultiUAV-135 includes an occluded UAV (red box, ID: 29) and a flying creature misclassified as a UAV (pink box, ID: 28); MultiUAV-173 features a complex background, where IDs 16, 17, and 18 are misjudgments; and MultiUAV-261 presents nearly invisible UAVs, leading to missed detections and tracking failures. The last row presents heatmaps highlighting the model's difficulty in UAV perception, especially in MultiUAV-261.}
    \label{fig:predict}
\end{figure*}

\subsection{Leaderboard Results}
\label{ssec:leaderboard_results}

Based on all trials across the three tracks, as summarized in \cref{tab:results_track1_track2} and \cref{tab:results_track3}, we report the leaderboard results in \cref{tab:leaderboard}, which includes the top three scores for each track, our submitted scores, and the official baseline scores. While there remains a gap between our scores and the top three, 0.1332, 0.1971, and 0.0502 for Tracks 1, 2, and 3, respectively, our performance shows substantial improvement over the baselines. Specifically, we achieve approximately a twofold increase over the baseline scores in Tracks 1 and 3 and nearly a fivefold improvement in Track 2. Notably, these results were obtained without employing image enhancement techniques or leveraging temporal information during training. Integrating such advanced techniques from our strong baseline could significantly improve performance and make reaching a top-three position highly feasible.

\begin{table}
    \centering
    \begin{adjustbox}{max width=\linewidth}
    \begin{tabular}{@{}lccc@{}}
        \toprule
        Teams & Track 1 & Track 2 & Track 3 \\
        \midrule
         1st Place Team & 0.7323 & 0.6676 & 0.8499 \\
         2nd Place Team & 0.7308 & 0.5712 & 0.8132 \\
         3rd Place Team & 0.7145 & 0.5530 & 0.8111 \\
         \textbf{Strong Baseline} (ours) & 0.5813 & 0.3559 & 0.7609 \\
         The\_4th\_Anti\_UAV\_Baseline & 0.2965 & 0.0745 & 0.3747 \\
        \bottomrule
    \end{tabular}
    \end{adjustbox}
    \caption{Leaderboard results for the top three teams, our approach, and the official baseline across all three tracks. Our method achieved scores of 0.5813 (19th), 0.3559 (14th), and 0.7609 (5th) in Tracks 1, 2, and 3, respectively, while the official baseline scored 0.2965 (32nd), 0.0745 (20th), and 0.3747 (20th).}
    \label{tab:leaderboard}
\end{table}

\subsection{Discussion and Enhancement Techniques}
\label{ssec:discussion}

\begin{figure*}[t]
    \centering
    \includegraphics[width=\linewidth]{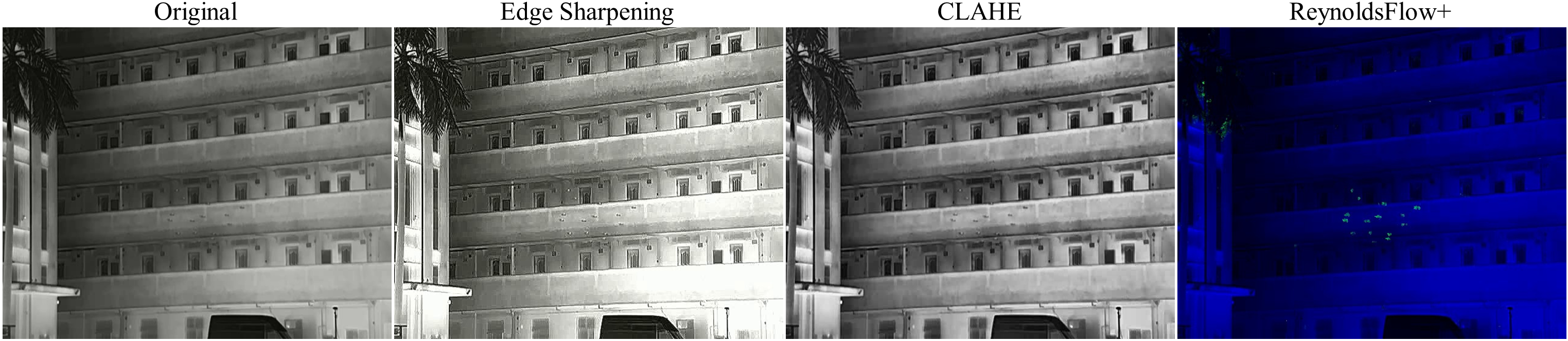}
    \caption{Potential frame enhancement techniques for multi-UAV tracking on a MultiUAV-262 video frame. From left to right: (1) original thermal infrared frame, (2) Sobel-based edge sharpening~\cite{gupta2016combining}, (3) contrast enhancement via Contrast Limited Adaptive Histogram Equalization (CLAHE)~\cite{musa2018review}, and (4) ReynoldsFlow+ visualization highlighting motion patterns to assist UAV detection~\cite{chen2025reynoldsflow}.}
    \label{fig:enhancement}
\end{figure*}

The evaluation results reveal several key insights. First, overfitting emerged due to our data-splitting strategy. To maximize scene diversity, we did not categorize videos by attributes such as fixed-camera setups or background types (\eg, sky or buildings). Instead, we directly split the dataset into training, validation, and testing sets, occasionally allowing frames from the same video to appear across splits. This likely contributed to overfitting, as evidenced by AP score discrepancies during local testing. Second, accurately rescaling the provided initial object positions to match the resolution used in training and inference is critical, as mismatches can mislead the tracker and degrade subsequent predictions. Third, increasing image resolution is key to breaking performance plateaus when parameter tuning fails to improve accuracy. For example, scaling from 640 to 1280 resolution yielded a significant score improvement of approximately 0.1. However, further increases produced diminishing gains as training at 2560 pixels for 7 epochs reached a score of 0.7072, and training at 3840 pixels for 1 epoch reached 0.7098, while both required significantly higher computational costs compared to training at 1280 pixels. Fourth, memory consumption during inference with YOLOv12 and BoT-SORT-ReID accumulates over time, leading to program crashes. To address this, we executed inference on a per-folder basis rather than processing all sequences in a single run. Finally, a clear performance gap is observed between runs with accurate initial object positions and those without, as evidenced by the performance difference between Tracks 1 and 2. This highlights the critical importance of promptly and reliably estimating initial positions to enhance tracking performance further.

Additionally, as previously discussed, while our approach provides a strong baseline, it remains insufficient for achieving top-tier performance without further refinement. \cref{fig:predict} (a) displays our model's predictions across various scenarios, while \cref{fig:predict} (b) highlights several key failure cases: (1) overlapping UAVs frequently cause ID switches, (2) distinguishing UAVs from flying creatures remains challenging, with the model often reassigning new IDs to UAVs following brief occlusions, (3) complex backgrounds lead to missed detections and tracking failures, and (4) tiny UAVs in cluttered environments provide little to no valuable information, making detection highly unreliable. The corresponding heatmaps in the last row illustrate the model's inability to perceive UAVs effectively in these challenging conditions. These limitations emphasize the importance of image enhancement techniques to improve performance further. \cref{fig:enhancement} illustrates several potential image enhancement methods. From left to right: (1) the original thermal infrared frame, (2) Sobel-based edge sharpening~\cite{gupta2016combining}, which highlights edges more clearly than the original, (3) Contrast Limited Adaptive Histogram Equalization (CLAHE)~\cite{musa2018review}, which improves contrast, and (4) ReynoldsFlow+~\cite{chen2025reynoldsflow}, a temporal enhancement method based on the Reynolds Transport Theorem~\cite{potter1975fluid}, a three-dimensional generalization of the Leibniz integral rule~\cite{flanders1973differentiation}, providing enhanced appearance for moving UAVs.
\section{Conclusion}
\label{sec:conclusion}

This paper presents a strong baseline for thermal infrared video-based multi-UAV tracking tasks. By integrating YOLOv12 with BoT-SORT, our approach significantly improves over the baseline. With additional strategies during training and inference, as discussed in the experimental results, we show that our method has the potential to rank in the top three, as seen in the Track 3 performance. We also identify key factors influencing performance compared to our initial trial: model size contributing approximately 0.001, track buffer size affecting the score by around 0.0001, image input size providing the most significant impact with a score increase of about 0.1, and the ReID module adding roughly 0.01. While our approach is intuitive and straightforward, we propose several potential techniques for further improving accuracy. Overall, our method establishes a strong baseline, primarily driven by the latest YOLOv12 detector and the advanced BoT-SORT tracking algorithm, making a strong starting point in recent advancements in the UAV swarm tracking field.
\section{Acknowledgments}
\label{sec:acknowledgements}

We thank the HPC system~\cite{meade2017spartan} at The University of Melbourne for providing the computational resources that significantly accelerated model training and enabled the completion of this paper.
{
    \small
    \bibliographystyle{ieeenat_fullname}
    \bibliography{main}
}


\end{document}